\pdfoutput=1
\documentclass[10pt,twocolumn,letterpaper]{article}

\usepackage[pagenumbers]{cvpr} 

\usepackage{graphicx}
\usepackage{amsmath}
\usepackage{amssymb}
\usepackage{booktabs}

\usepackage{multirow}

\usepackage[pagebackref,breaklinks,colorlinks]{hyperref}

\usepackage[capitalize]{cleveref}
\crefname{section}{Sec.}{Secs.}
\Crefname{section}{Section}{Sections}
\Crefname{table}{Table}{Tables}
\crefname{table}{Tab.}{Tabs.}

\def\cvprPaperID{10734} 
\def\confName{CVPR}
\def\confYear{2023}

\begin{document}

\title{Hierarchical Semantic Contrast for Scene-aware Video Anomaly Detection}

\author{Shengyang Sun \quad\quad\quad\quad\quad Xiaojin Gong\thanks{Corresponding author.}\\
College of Information Science \& Electronic Engineering,\\
Zhejiang University, Hangzhou, Zhejiang, China\\
{\tt\small \{sunshy,gongxj\}@zju.edu.cn}
}
\maketitle
\begin{abstract}
Increasing scene-awareness is a key challenge in video anomaly detection (VAD). In this work, we propose a hierarchical semantic contrast (HSC) method to learn a scene-aware VAD model from normal videos. We first incorporate foreground object and background scene features with high-level semantics by taking advantage of pre-trained video parsing models. Then, building upon the autoencoder-based reconstruction framework, we introduce both scene-level and object-level contrastive learning to enforce the encoded latent features to be compact within the same semantic classes while being separable across different classes. This hierarchical semantic contrast strategy helps to deal with the diversity of normal patterns and also increases their discrimination ability. Moreover, for the sake of tackling rare normal activities, we design a skeleton-based motion augmentation to increase samples and refine the model further. Extensive experiments on three public datasets and scene-dependent mixture datasets validate the effectiveness of our proposed method. 
\end{abstract}

\section{Introduction}
\label{sec:intro}
With the prevalence of surveillance cameras deployed in public places, video anomaly detection (VAD) has attracted considerable attention from both academia and industry. It aims to automatically detect abnormal events so that the workload of human monitors can be greatly reduced. By now, numerous VAD methods have been developed under different supervision settings, including weakly supervised~\cite{sultani2018real,ijcai2021p162,tian2021weakly,feng2021mist,zhong2019graph,sun2023LSTC}, purely unsupervised~\cite{zaheer2022generative,Yu2022unsupervised}, and ones learning from normal videos only~\cite{hasan2016learning,liu2018future,park2020learning,ionescu2019object,ramachandra2020learning}. However, it is extremely difficult or even impossible to collect sufficient and comprehensive abnormal data due to the rare occurrence of anomalies, whereas collecting abundant normal data is relatively easy. Therefore, the setting of learning from normal data is more practical and plays the dominant role in past studies.

\begin{figure}[t]
	\centering
	\includegraphics[width=0.99\linewidth]{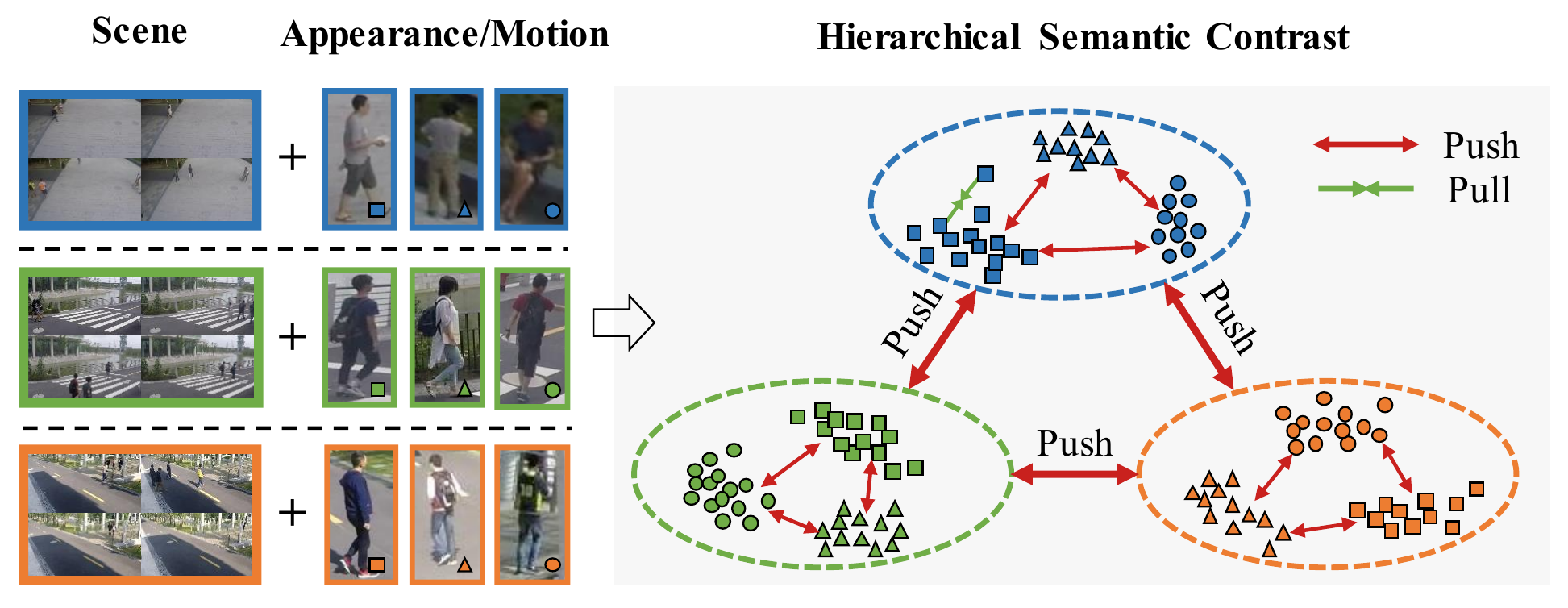}
	\caption{An illustration of hierarchical semantic contrast. The encoded scene-appearance/motion features are gathered together with respect to their semantic classes. Best viewed in color.}
	\label{fig:index}
	\vspace{-8pt}
\end{figure}

Although a majority of previous techniques learn their VAD models from normal data, this task has still not been well addressed due to the following reasons. First, some anomalies are scene-dependent~\cite{9271895,sun2020scene}, implying that an appearance or motion may be anomalous in one scene but normal in other scenes. How to detect scene-dependent anomalies while preventing background bias (\ie learning the background noise rather than the essence of anomaly~\cite{liu2019exploring}) is a challenging problem. Second, normal patterns are diverse. How to enable a deep VAD model to represent the diverse normality well but not generalize to anomalous data is also a challenge~\cite{gong2019memorizing,park2020learning}. Last but not least, samples collected from different normal patterns are imbalanced because some normal activities may appear very sparsely~\cite{9271895}. How to deal with rare but normal activities is challenging as well. 

Previous VAD methods mainly perform learning at frame-level~\cite{hasan2016learning,zhao2017spatio,Ravanbakhsh2019gan} or in an object-centric~\cite{ionescu2019object,Georgescu2022AED,Zhou2022object} way. The former is prone to suffer from the background bias~\cite{liu2019exploring} while most of the latter methods are background-agnostic. There are some attempts to address the above-mentioned challenges in one or another aspect. For instance, a spatio-temporal context graph~\cite{sun2020scene} and a hierarchical scene normality-binding model~\cite{bao2022scene} are constructed to discover scene-dependent anomalies. Memory-augmented autoencoders (AE)~\cite{gong2019memorizing,park2020learning} are designed to represent diverse normal patterns while lessening the powerful capacity of AEs. An over-sampling strategy~\cite{Liu2019imbalance} is adopted but to solve the imbalance between normal and abnormal data. Contrastively, in this work we address all of these challenges simultaneously and in distinct ways.

The primary objective of our work is to handle scene-dependent anomalies. An intuition behind scene-dependent anomalies is that, if a type of object or activity is never observed in one scene in normal videos, then it should be viewed as an anomaly. It implies that we can first determine the scene type and then check if an object or activity has occurred in normal patterns of this scene. Based on this observation, we propose a hierarchical semantic contrast method to learn a scene-aware VAD model. Taking advantage of pre-trained video parsing networks, we group the appearance and activity of objects and background scenes into semantic categories. Then, building upon the autoencoder-based reconstruction framework, we design both scene-level and object-level contrastive learning to enforce the encoded latent features to gather together with respect to their semantic categories, as shown in Fig.~\ref{fig:index}. When a test video is input, we retrieve weighted normal features for reconstruction and the clips of high errors are detected as anomalies. 

The contributions of this work are as follows:
\vspace{-3pt}
\begin{itemize}
	\item We build a scene-aware reconstruction framework composed of scene-aware feature encoders and object-centric feature decoders for anomaly detection. The scene-aware encoders take background scenes into account while the object-centric decoders are to reduce the background noise.
	\vspace{-2pt}
	\item We propose hierarchical semantic contrastive learning to regularize the encoded features in the latent spaces, making normal features more compact within the same semantic classes and separable between different classes. Consequently, it helps to discriminate anomalies from normal patterns.
	\vspace{-2pt}
	\item We design a skeleton-based augmentation method to generate both normal and abnormal samples based on our scene-aware VAD framework. The augmented samples enable us to additionally train a binary classifier that helps to boost the performance further. 
	\vspace{-2pt}
	\item {Experiments on three public datasets demonstrate promising results on scene-independent VAD. Moreover, our method also shows a strong ability in detecting scene-dependent anomalies on self-built datasets.}
	\vspace{-2pt}
\end{itemize}

\section{Related Work}
\subsection{Video Anomaly Detection}
Most previous VAD studies can be grouped into weakly supervised category~\cite{sultani2018real,ijcai2021p162,tian2021weakly,feng2021mist,zhong2019graph,sun2023LSTC} that learns with video-level labels, or the one learning from normal videos only~\cite{hasan2016learning,zhao2017spatio,gong2019memorizing,park2020learning,Ravanbakhsh2019gan}. In this work, we focus on the latter category, which is mainly addressed by reconstruction- or distance-based techniques. The reconstruction-based techniques use autoencoder (AE)~\cite{hasan2016learning,zhao2017spatio,lv2021learning}, memory-augmented AE~\cite{gong2019memorizing,park2020learning,liu2021hybrid}, or generative models~\cite{Ravanbakhsh2019gan,Nguyen2020GAN} to reconstruct current frame~\cite{hasan2016learning,zhao2017spatio,gong2019memorizing,park2020learning} or predict future frames~\cite{liu2018future,Nguyen2020GAN}, by which the frames of high reconstruction errors are detected as anomalies. The distance-based techniques often adopt one-class SVMs~\cite{ionescu2019object,ionescu2019detecting}, Gaussian mixture models~\cite{Sabokrou2017Gaussian,Sun2022reasoning}, or other classifiers~\cite{Georgescu2022AED} to compute a decision boundary and those deviating from the normality are screened out as anomalies.

A majority of reconstruction- and distance-based techniques~\cite{hasan2016learning,zhao2017spatio,gong2019memorizing,park2020learning,Ravanbakhsh2019gan,Nguyen2020GAN,ionescu2019detecting} learn their models at frame-level, which may suffer from the background bias~\cite{liu2019exploring} and lack of explainability. To this end, various object-centric methods have been developed, leveraging appearance and motion~\cite{ionescu2019object,Georgescu2022AED,Zhou2022object,georgescu2021anomaly,Yang2022twostream}, or skeleton~\cite{morais2019learning,markovitz2020graph,li2022self,Yang2022twostream} of objects to promote the performance. However, the VAD models learned by most of them are background-agnostic. Considering that some anomaly events are scene-dependent, a few scene-aware methods~\cite{sun2020scene,bao2022scene,Sun2022reasoning,Cao2022context} have been proposed recently. For instance, Sun \etal~\cite{sun2020scene} construct a spatio-temporal context graph to represent both objects and the background, Sun \etal~\cite{Sun2022reasoning} and Bao \etal~\cite{bao2022scene} learn memory-augmented AEs to encode scene and objects, and Cao \etal~\cite{Cao2022context} design a network with context recovery and knowledge retrieval streams. Our work adopts the autoencoder-based reconstruction framework like~\cite{bao2022scene}. But differently, we build scene-aware encoders and object-centric decoders for reconstruction and propose the hierarchical semantic contrast to regularize the encoded latent features.

\begin{figure*}[t]
	\centering
	\includegraphics[width=0.99\textwidth]{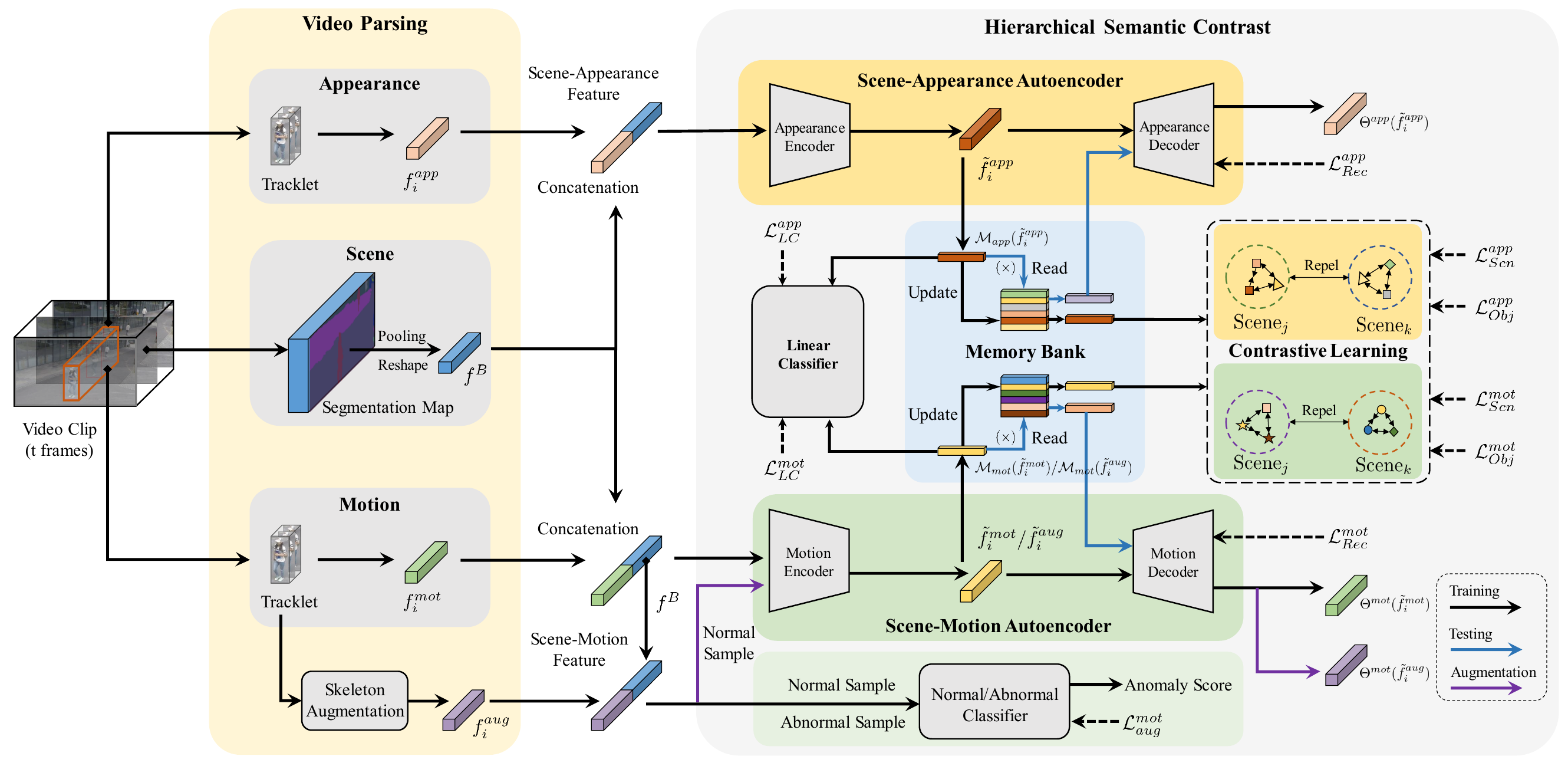} 
	\caption{An overview of the proposed method. It consists of video parsing, scene-aware autoencoders, memory-based contrastive learning, and motion augmentation modules. Best viewed in color.}	
	\label{fig:framework}
	\vspace{-12pt}
\end{figure*}

\subsection{Contrastive Learning}
Contrastive learning has been successfully applied to various vision tasks, such as representation learning~\cite{wu2018memory,Chen2020SimCLR,he2020momentum,Li2021pcl}, person re-identification~\cite{ge2020self,Wang2022O2CAP}, and semantic segmentation~\cite{Du2022seg,Zhou2022seg}. It performs learning via contrasting anchor instances with their positive and negative instances or prototypes, which are sampled either from a large batch~\cite{Chen2020SimCLR} or from an external memory bank~\cite{he2020momentum,Wang2022O2CAP}. Recently, contrastive learning has also been exploited in anomaly detection~\cite{Okan2021CL_DAD,Wang2022CL_AD,wang2020cluster,Huang2022contr,Lu2022contr}. Most methods~\cite{Okan2021CL_DAD,wang2020cluster,Huang2022contr,Lu2022contr} perform contrast between an instance and its augmented version, focusing on the instance level only. An exceptional work is HSCL~\cite{Wang2022CL_AD}, which takes into account sample-to-sample, sample-to-prototype, and normal-to-abnormal contrasts to implement semi-supervised anomaly detection. In our work, we design a hierarchical contrastive learning strategy that performs contrast at the scene-level and object-level, enforcing instances to gather together according to their semantic categories.

\subsection{Data Augmentation}
Data augmentation is extensively used in contrastive learning and other class-imbalanced learning tasks~\cite{lim2018doping}. The works most related to ours are skeleton augmentation methods. For instance, Meng \etal~\cite{Meng2021aug} design a transformation network to generate new skeleton samples. Thoker \etal~\cite{Thoker2021skeleton} design spatial and temporal skeleton augmentation based on shear transformation and joint jittering. Guo \etal~\cite{Guo2022skeleton} apply shear, spatial flip, rotate, and axis mask to generate extreme augmentations. These methods apply skeleton-based augmentation to generate positive samples for the action recognition task. In contrast, we design a skeleton-based augmentation to produce both normal and abnormal samples of rare activities, helping the learning of imbalanced anomaly detection.  

\section{The Proposed Method}
Figure~\ref{fig:framework} presents an overview of the proposed method. When a video clip (\ie a set of consecutive frames) is input, we first parse it to get high-level semantic features, including the appearance and motion of objects, together with the background scene. Then, the appearance or motion feature of each object is incorporated with the scene feature. The obtained scene-appearance and scene-motion features are fed into scene-aware encoders and object-centric decoders for feature encoding and reconstruction. All encoded latent features are stored in external memory banks, based on which we perform scene- and object-level semantic contrastive learning. The hierarchical contrastive learning enforces the diverse latent normal features to be compact within the same semantic classes and separable between different classes, which consequently increases the discrimination ability of normal patterns. During inference time, normal features stored in memory are retrieved and weighted to reconstruct the features of objects in a test clip, and those with high errors are detected as anomalies. 

\subsection{Video Parsing}
Pre-trained video parsing networks are extensively used in many VAD methods~\cite{ionescu2019object,Zhou2022object,georgescu2021anomaly,Yang2022twostream,morais2019learning,markovitz2020graph,Yang2022twostream,sun2020scene,bao2022scene,Sun2022reasoning} to extract different visual cues. In this work, we take advantage of several pre-trained networks to extract high-level features while introducing semantic labels. 

Given a video clip $\mathcal{C}$ composed of $T$ consecutive frames, we first adopt the pre-trained YOLOv3~\cite{redmon2018yolov3} and FairMOT~\cite{zhang2021fairmot} to detect and track objects, which produce several object tracklets and their semantic class labels such as \textit{pedestrian}, \textit{bicycle}, \etc. Then, we extract both appearance and motion features for each object tracklet and extract a scene feature for the remaining background as follows.

\noindent\textbf{Appearance feature extraction.} Appearance information plays an important role in detecting appearance anomalies. Therefore, for an object tracklet $\mathcal{O}_{i}$ in the clip, we employ ViT~\cite{dosovitskiy2020image} to extract an appearance feature for each frame of the tracklet, and the features of all frames are averaged to generate one appearance feature $f_{i}^{app}\in\mathbb{R}^{1024}$.

\noindent\textbf{Motion feature extraction.} Motion information is of equal importance in VAD. Considering that human-related anomalies are dominant in non-traffic surveillance, we opt to extract action information as a motion feature instead of using optical flow. More specifically, for an object tracklet $\mathcal{O}_{i}$, we use a pre-trained HRNet~\cite{sun2019deep} to extract a skeleton feature for each frame. The features of all frames are further fed into PoseConv3D~\cite{duan2022revisiting} to produce one motion feature $f_{i}^{mot}\in\mathbb{R}^{512}$, together with an action class label such as \textit{walking}, \textit{jumping}, \textit{kicking}, \etc.

\noindent\textbf{Scene feature extraction.} In pursuit of scene-awareness, we also extract a scene feature for the clip background. For each clip frame, we employ DeepLabV3+~\cite{chen2018encoder} to generate a segmentation map while masking out the foreground object categories. Then, we perform max-pooling, reshape, averaging, and $l_2$ normalization on all segmentation maps to obtain one scene feature $f^{B}\in\mathbb{R}^{D_B}$, where $D_B$ depends on the size of the video frame. To discriminate different scenes at a fine-grained level, we utilize DBSCAN~\cite{ester1996density} for clustering and generating pseudo labels of scene classes.  

\subsection{Semantic Feature Reconstruction}
In this work, we adopt the extensively used reconstruction framework for our anomaly detection. For each appearance or motion feature, we design an autoencoder composed of a scene-aware encoder and an object-centric decoder for feature reconstruction. 

\noindent\textbf{Scene-aware feature encoder.} To correlate foreground objects with the background scene, we incorporate each appearance/motion feature with its corresponding scene feature. The obtained scene-appearance or scene-motion feature is fed into a scene-aware feature encoder. Formally, it is represented by
\begin{equation} \label{eq:embedding}
	\tilde{f}^{*}_{i} = \Phi^{*}([f^{B}, f^{*}_{i}]),
\end{equation}
where $\tilde{f}^{*}_{i}\in\mathbb{R}^{D_E}$ is the encoded latent feature of object $\mathcal{O}_{i}$ in clip $\mathcal{C}$, `$*$' denotes either $app$ or $mot$, and $D_E$ is the feature dimension. Moreover, $[\cdot, \cdot]$ denotes the concatenation and $\Phi^{*}(\cdot)$ is the feature encoder, which is implemented by a two-layer MLP followed with a $l_2$ normalization.

\noindent\textbf{Object-centric feature decoder.} The reconstruction-based framework assumes that anomalies cannot be represented well by normal patterns. To reduce the background bias~\cite{liu2019exploring} in reconstruction, we opt to reconstruct the feature of each foreground object instead of the incorporated scene-aware feature. That is, given a latent code $\tilde{f}^{*}_{i}$, we enforce the decoder to reconstruct a feature close to the appearance/motion feature $f^{*}_{i}$, which is
\begin{equation} \label{eq:loss_Recon}
	\mathcal{L}^*_{Rec} =
	\lVert f^{*}_{i} - \Theta^{*}(\tilde{f}^{*}_{i})\rVert^2_2,
\end{equation}
where $\lVert\cdot\rVert_2$ is the $l_2$ norm. $\Theta^{*}$ is the feature decoder implemented by a two-layer MLP as well.

\subsection{Hierarchical Semantic Contrast}
Due to the diversity of normal patterns as well as the large capacity of deep networks, the model learned from normal data may also reconstruct anomalies well~\cite{gong2019memorizing,park2020learning}. To address this problem, we propose a hierarchical semantic contrast (HSC) strategy to regularize the encoded normal features in the latent space, by which diverse normal patterns can be represented more compactly and therefore be more discriminative to anomalies. HSC conducts contrastive learning at the scene- and object-level by taking advantage of the semantic labels introduced in video parsing. 

\noindent\textbf{Scene-level contrastive learning.}
The scene-level contrastive learning aims to attract the latent features within the same scene class and repel the features of different scenes. To this end, we adopt the InfoNCE loss~\cite{wu2018memory,Chen2020SimCLR} to conduct learning, assisted by an external memory bank. The scene-level contrastive loss is defined by
\begin{equation} \label{eq:loss_inter_class}
	\mathcal{L}^*_{Scn}= - \sum\limits_{\tilde{f}^{*}_j\in\mathcal{X}_*(\tilde{f}^{*}_i)}
	log\frac{exp(sim(\tilde{f}^{*}_i, \tilde{f}^{*}_j)/\tau)}{\sum\nolimits^{N}_{k=1}exp(sim(\tilde{f}^{*}_i, \tilde{f}^{*}_k)/\tau)},
\end{equation}
where $N$ is the number of all encoded latent features, $\mathcal{X}_*(\tilde{f}^{*}_i)$ indicates the set of features sharing the same pseudo scene label with $\tilde{f}^{*}_i$, $\tau$ is the temperature hyperparameter, and $sim(\cdot,\cdot)$ denotes the cosine similarity.

Besides, we also build a linear classification (LC) head to classify each latent feature into its pseudo scene class by using the cross-entropy loss:
\begin{equation} \label{eq:loss_LC}
	\mathcal{L}^*_{LC} = - log <\mathcal{Y}(\tilde{f}^{*}_i),\Lambda^{*}(\tilde{f}^{*}_i)>, 
\end{equation}
where $<\cdot,\cdot>$ denotes dot product, $\Lambda^{*}(\cdot)$ is the linear classifier, and $\mathcal{Y}$ represents the pseudo scene label of $\tilde{f}^{*}_i$.

\noindent\textbf{Object-level contrastive learning.}
Within each scene class, the object-level contrastive learning pulls the latent features of the same appearance/motion category together and pushes away those from different appearance/motion categories. Therefore, the object-level contrastive loss is defined by
\begin{equation} \label{eq:loss_intra_class}
	\mathcal{L}^*_{Obj}=-
	\sum\limits_{\tilde{f}^{*}_j\in\mathcal{N}_*(\tilde{f}^{*}_i)}
	log\frac{exp(sim(\tilde{f}^{*}_i, \tilde{f}^{*}_j)/\tau)}
	{\sum\nolimits_{\tilde{f}^{*}_k\in\mathcal{X}_*(\tilde{f}^{*}_i)}
		exp(sim(\tilde{f}^{*}_i, \tilde{f}^{*}_k)/\tau)},
\end{equation}
where $\mathcal{N}_*(\tilde{f}^{*}_i)$ represents the set of latent features sharing the same appearance/motion class and same scene class with $\tilde{f}^{*}_i$. Note that in this loss only the features within the same scene class are considered and all others are ignored. 

\noindent\textbf{Memory banks.} In contrast to the memory-augmented AEs~\cite{gong2019memorizing,park2020learning} that utilize memory for the learning of autoencoders, we use memory mainly for our contrastive learning. To this end, two memory banks are built for storing the latent scene-appearance and scene-motion features respectively. Each entry is updated by 
\begin{equation} \label{eq:memory_update}
	\mathcal{M}_*(\tilde{f}^{*}_i)\gets (1-m)\tilde{f}^{*}_i+m\mathcal{M}_*(\tilde{f}^{*}_i),
\end{equation}
followed with a $l_2$ normalization, where $m\in[0,1)$ is a momentum coefficient.

\subsection{Motion Augmentation}
\begin{figure}[t]
	\centering
	\includegraphics[width=1\linewidth]{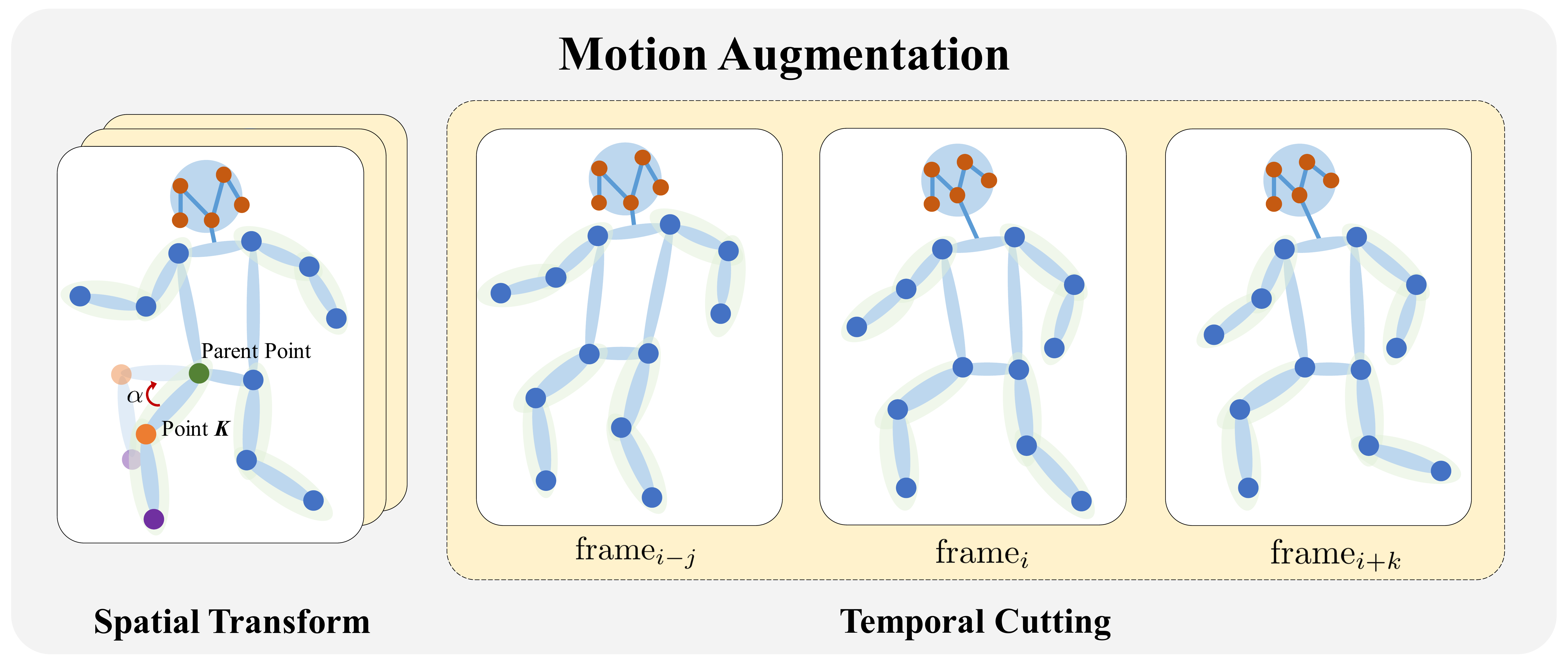}
	\caption{An illustration of our skeleton-based motion augmentation, which consists of spatial transformation and temporal cutting.}
	\label{fig:onecol1}
\end{figure}

The occurrence of rare but normal activities is a challenge in VAD~\cite{9271895}. This challenge stands out in scene-dependent anomaly detection when compared to the scene-agnostic case. The reason is that normal samples collected from different scenes are not counted together anymore. To address this problem, we design a skeleton-based augmentation to produce more samples, which includes spatial transformation and temporal cutting as shown in Fig.~\ref{fig:onecol1}.

\noindent\textbf{Spatial transformation.} A skeleton feature extracted from one object frame contains a set of human anatomical keypoints including \textit{shoulder}, \textit{elbow}, \textit{wrist}, \etc. In this work, we design a rotation-based augmentation scheme. For each keypoint $K$ except those on \textit{head}, we set a probability $P_{st}$ to decide if the keypoint is rotated or not. If the keypoint $K$ is chosen to rotate, it rotates around its parent node and the new coordinates $K_{rot}$ are obtained by
\begin{equation} \label{eq:spatial_transformation}
K_{rot} = (K - P(K))\begin{bmatrix}
		 cos(\alpha) & sin(\alpha) \\
		 -sin(\alpha) & cos(\alpha)
		\end{bmatrix} + P(K),
\end{equation}
where $P(K)$ is the parent keypoint of $K$, and $\alpha$ is a rotation angle randomly selected within a pre-defined range. Moreover, when $K$ is rotated, its descendant keypoints are all rotated consequently. 

\noindent\textbf{Temporal cutting.} An action is identified not only by the spatial distribution of keypoints but also by the temporal distribution. In this work, we simply adopt the cutting strategy for temporal augmentation. That is, given the frames of an object tracklet, we set a probability $P_{tc}$ for each frame to decide if it is left out or not.

\noindent\textbf{Spatio-temporal augmentation.} To increase the diversity of motion samples, we combine spatial transformation and temporal cutting together as our spatio-temporal augmentation. Given an object tracklet, we apply the spatio-temporal augmentation to produce a new set of skeleton features and then feed them into PoseConv3D~\cite{duan2022revisiting} to obtain the motion feature of the augmented sample.

\subsection{Training and Test}

\noindent\textbf{Training.} The training loss of our full model contains a loss $\mathcal{L}^{app}$ for the appearance stream and a loss $\mathcal{L}^{mot}$ for the motion stream. That is, the total loss is defined by
\begin{equation} \label{eq:model_loss}
	\mathcal{L}=\mathcal{L}^{app}+\mathcal{L}^{mot}.
\end{equation}
Here, the loss for each stream consists of two contrastive losses, together with a classification loss and a reconstruction loss. That is,
\begin{equation} \label{eq:branch_loss}
\mathcal{L}^{*}=\mathcal{L}^*_{Scn}+\mathcal{L}^*_{Obj}+\mathcal{L}^*_{LC}+\mathcal{L}^*_{Rec},
\end{equation}
in which $*$ denotes either $app$ or $mot$ as before.

At the first stage of training, we use the loss $\mathcal{L}$ to train our model on an original dataset without motion augmentation. Once the model is trained, we take augmented samples into consideration for refinement. Since the samples generated in motion augmentation are not guaranteed to be normal, we apply our trained model to discriminate normal and abnormal samples based on their reconstruction errors defined in \cref{eq:branch_anomaly_score}. Then, we leverage both normal and abnormal samples to additionally train a binary classifier on the motion stream using a cross-entropy loss $\mathcal{L}^{mot}_{aug}$.

\noindent\textbf{Test.} During inference time, we apply video parsing to obtain high-level features for each test video clip. Then, each test feature $f^{*}_t$ is fed into the appearance/motion stream for encoding and reconstruction. Let us denote the encoded latent feature as $\tilde{f}^{*}_t$. Different from training that directly reconstructs the latent feature, we calculate the similarity between it to each entry stored in the memory $\mathcal{M}_*$ by 
\begin{equation} \label{eq:memory_matching}
	w_i=\frac{exp( 
		(\tilde{f}^{*}_t)^{T}\mathcal{M}_*(i))}
	    {\sum\nolimits^{N}_{i=1}
	    exp((\tilde{f}^{*}_t)^{T}\mathcal{M}_*(i))},
\end{equation}
and get a weighted average of all stored normal features for reconstruction. 

The reconstruction error of one stream is therefore defined by
\begin{equation} \label{eq:branch_anomaly_score}
	\mathcal{S}^{*}(f^{*}_t) =
	\lVert f^{*}_t - \Theta^*(\sum\limits^{N}_{i=1}
	w_i\mathcal{M}_*(i))\rVert^2_2.
\end{equation}
The final anomaly score of an object is defined as the average reconstruction error of two streams, which is 
\begin{equation} \label{eq:final_anomaly_score}
	\mathcal{S}(f^{app}_t, f^{mot}_t) = 
	\frac{1}{2}(\mathcal{S}^{app}(f^{app}_t)+\mathcal{S}^{mot}(f^{mot}_t)).
\end{equation}
When motion augmentation is considered, the anomaly score of the motion stream is replaced by the anomaly probability output by the binary classifier. Moreover, the anomaly score of a clip is decided by the highest final anomaly score of objects in this clip. Finally, we apply a Gaussian filter for temporal smoothing over all video clips.

\section{Experiments}
\subsection{Datasets and Evaluation Metrics}
We evaluate the proposed method on three public datasets: UCSD Ped2~\cite{li2013anomaly}, Avenue~\cite{Lu2013avenue}, and ShanghaiTech~\cite{liu2018future}. UCSD Ped2~\cite{li2013anomaly} is a single-scene dataset collected from pedestrian walkways, including anomalies such as \textit{bikers}, \textit{skaters}, small \textit{carts} across a walkway. Avenue~\cite{Lu2013avenue} is a single-scene dataset as well. It is captured in CUHK campus avenue, containing anomalies like \textit{running}, \textit{bicycling}, \etc. It also contains some rare normal patterns~\cite{Lu2013avenue}. ShanghaiTech~\cite{liu2018future} is a challenging multi-scene dataset containing 13 campus scenes with various light conditions and camera angles. The statistics of these datasets are summarized in Table~\ref{tab:datasets}. 

However, these three datasets contain very few scene-dependent anomalies. And as far as we know, there is no public scene-dependent anomaly dataset available. In order to investigate the performance of our method on scene-dependent anomaly detection, we additionally create three mixture datasets based on ShanghaiTech. The mixture set $[01,02]$ consists of videos taken from scenes 01 and 02. We select a part of test videos of scene 01 containing the \textit{cyclist} events into the mixture training set and delete them from the test set. It implies that \textit{cyclist} is normal in scene 01, but it is still abnormal in scene 02. Likewise, we get a mixture set $[04,08]$ and a set $[10,12]$, in which some events are normal in one scene but abnormal in the other scene. More details are provided in our supplementary material.

For performance evaluation, we adopt the area under the curve (AUC) of the frame-level receiver operating characteristics (ROC) as the evaluation metric following the common practice~\cite{gong2019memorizing, wang2022ROADMAP, fang2022SIGnet, lv2021learning, cai2021appearance, Georgescu2022AED, chang2020clustering, ramachandra2020learning}. It concatenates all frames and then computes the score, also known as micro-averaged AUC~\cite{Georgescu2022AED}. 
 
 \begin{table}[b]
 	\vspace{-15pt}
 	\centering
 	\caption{The statistics of three public datasets and self-built scene-dependent datasets.}
 	\vspace{-3pt}
 	\resizebox{.47\textwidth}{!}{
 		\begin{tabular}{c|c|c|c|c}
 			\toprule
 			\multirow{2}*{Dataset} & Training & Test & \multirow{2}*{Scene} & \multirow{2}*{Resolution} \\
 			& Frame &  Frame &  & \\
 			\midrule
 			UCSD Ped2~\cite{li2013anomaly} & 2,550 & 2,010 & 1 & 360$\times$240 \\
 			CUHK Avenue~\cite{Lu2013avenue} & 15,328 & 15,324 & 1 & 640$\times$360 \\
 			ShanghaiTech~\cite{liu2018future} & 274,515 & 42,883 & 13 & 856$\times$480 \\
 			\midrule
 			Mixture $[01,02]$ & 14,080 & 5,488 & 2 & 856$\times$480  \\
 			Mixture $[04,08]$ & 37,600 & 5,104 & 2 & 856$\times$480  \\
 			Mixture $[10,12]$ & 33,856 & 3,584 & 2 & 856$\times$480  \\
 			\bottomrule
 	\end{tabular}}
 	\label{tab:datasets}
 \end{table}

\subsection{Implementation Details}
We implement the proposed method in Pytorch. The hyper-parameters involved in our model are set as follows. The dimension of encoded latent features is $D_E = 1280$. The temperature factor in contrastive learning is $\tau = 0.5$ and the momentum coefficient in memory updating is $m = 0.9$. The probabilities used for motion augmentation are set as $P_{st} = P_{tc} = 0.5$. In addition, our model is trained using the AdaGrad optimizer with a learning rate of 0.01 and a batch size of 128 for both UCSD Ped2 and Avenue and 512 for ShanghaiTech. Some other details are provided in our supplementary material. 

\subsection{Ablation Studies}
Although the proposed method aims at scene-dependent VAD, it works for scene-independent anomalies as well. Therefore, we conduct ablation studies mostly on Avenue and ShanghaiTech and partially on the mixture sets.

\begin{table}[tbp]
	\vspace{-5pt}
	\centering
	\caption{The AUC(\%) performance of our model variants.}
	\vspace{-5pt}
	\resizebox{.48\textwidth}{!}{
		\begin{tabular}{cccc|cc}
			\toprule
			MemCL & SA-AE & SM-AE & MA & Avenue & ShanghaiTech \\
			\midrule
			~ & \checkmark & ~ & ~ & 90.6 & 78.4 \\
			~ & ~ & \checkmark  & ~ & 81.3 & 77.6 \\
			~ & ~ & \checkmark & \checkmark & 82.6 & 77.8 \\
			~ & \checkmark & \checkmark & ~ & 91.1 & 80.7 \\
			~ & \checkmark & \checkmark & \checkmark & 91.5 & 81.2 \\
			\midrule
			\checkmark & \checkmark & ~ & ~ & 92.1 & 79.3 \\
			\checkmark & ~ & \checkmark  & ~ & 82.9 & 78.1 \\
			\checkmark & ~ & \checkmark & \checkmark & 84.9 & 78.3 \\
			\checkmark & \checkmark & \checkmark & ~ & 92.4 & 83.0 \\
			\checkmark & \checkmark & \checkmark & \checkmark & \textbf{93.7} & \textbf{83.4} \\
			\bottomrule
		\end{tabular}
	}
	\label{tab:ablation_model_components}
	\vspace{-12pt}
\end{table}

\label{sec:ablation}
\noindent\textbf{Effectiveness of the proposed components.} We first validate the effectiveness of our proposed components. We decompose the full model into scene-appearance autoencoder (SA-AE), scene-motion autoencoder (SM-AE), and memory-based contrastive learning (MemCL), together with scene-motion augmentation (MA) components. The performance of the model variants holding different components is reported in Table~\ref{tab:ablation_model_components}. From the results, we observe that SA-AE outperforms SM-AE or SM-AE+MA when only a single stream is learned and the combination of both streams performs better. Besides, memory-based contrastive learning enables the models to outperform their counterparts by a considerable margin. Motion augmentation also improves the performance on both datasets, especially on the Avenue dataset that contains rare normal activities.

\begin{figure*}[t]
	\centering
	\includegraphics[width=0.95\textwidth]{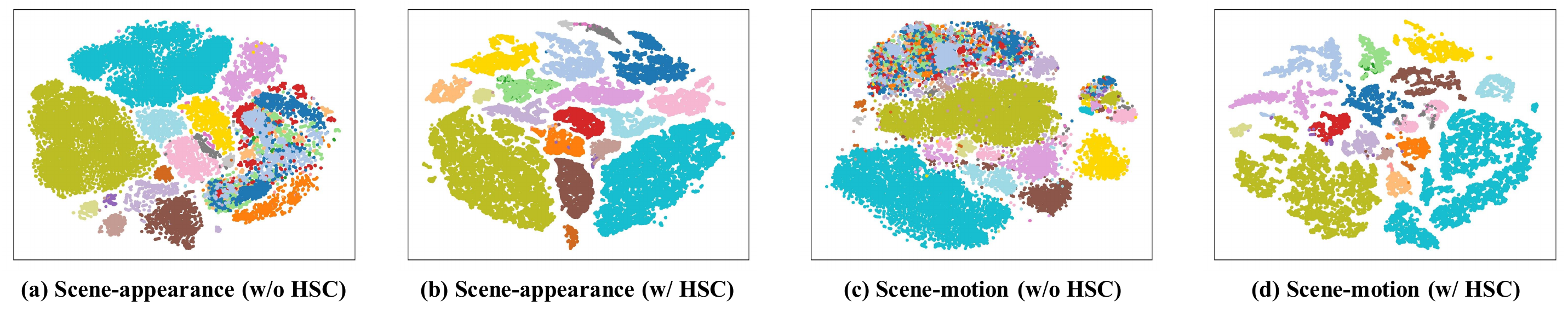} 
	\caption{t-SNE~\cite{van2008visualizing} visualization of the scene-appearance/motion features encoded by our models without or with hierarchical semantic contrast. The points with the same color belong to the identical scene. Best viewed in color.}
	\label{fig:visual_inter_clustering}
	\vspace{-10pt}
\end{figure*}

\noindent\textbf{Effectiveness of scene-aware AEs and HSC.} We here go deeper into the above-mentioned components for investigation. More specifically, we check the effectiveness of the scene-aware feature encoder (SA-E) and object-centric feature decoder (OC-D) in our autoencoders, together with the contrastive losses used in hierarchical semantic contrast (HSC). We conduct a series of experiments on the model without using motion augmentation. The results are presented in Table~\ref{tab:ablation_contrast_module}. It shows that, when contrastive learning is not applied, the scene-aware feature encoder slightly degenerates the performance on scene-independent Avenue and ShanghaiTech but improves the performance on the scene-dependent mixture sets. Moreover, the object-centric decoder improves the performance of all datasets since the background noise in reconstruction is avoided. In HSC, the individual contrastive learning at either scene- or object-level can consistently boost the performance, indicating the necessity of regularizing encoded features in the latent space. And the best performance is achieved when the losses work together.

\begin{figure}[t]
	\vspace{-4pt}
	\centering
	\includegraphics[width=1\linewidth]{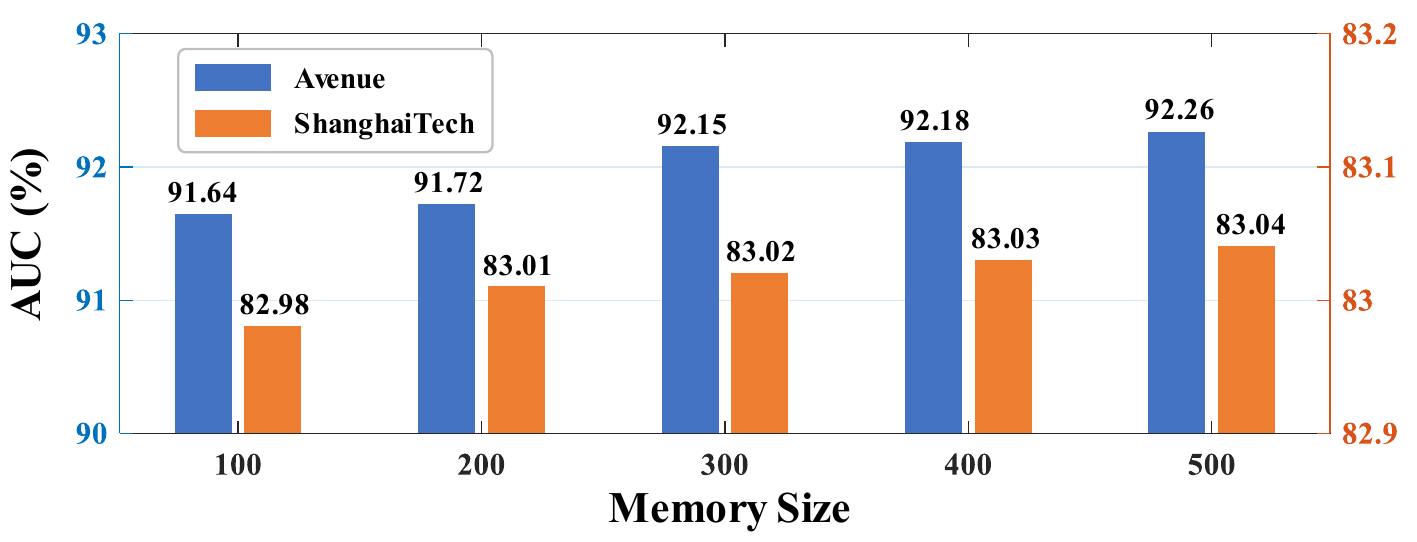}
	\vspace{-18pt}
	\caption{The AUC(\%) performance varies with respect to the memory size on Avenue and ShanghaiTech. Best viewed in color.}
	\label{fig:ablation_memory_size}
	\vspace{-10pt}
\end{figure}

\begin{figure}[t]
	\vspace{-12pt}
	\centering
	\includegraphics[width=1\linewidth]{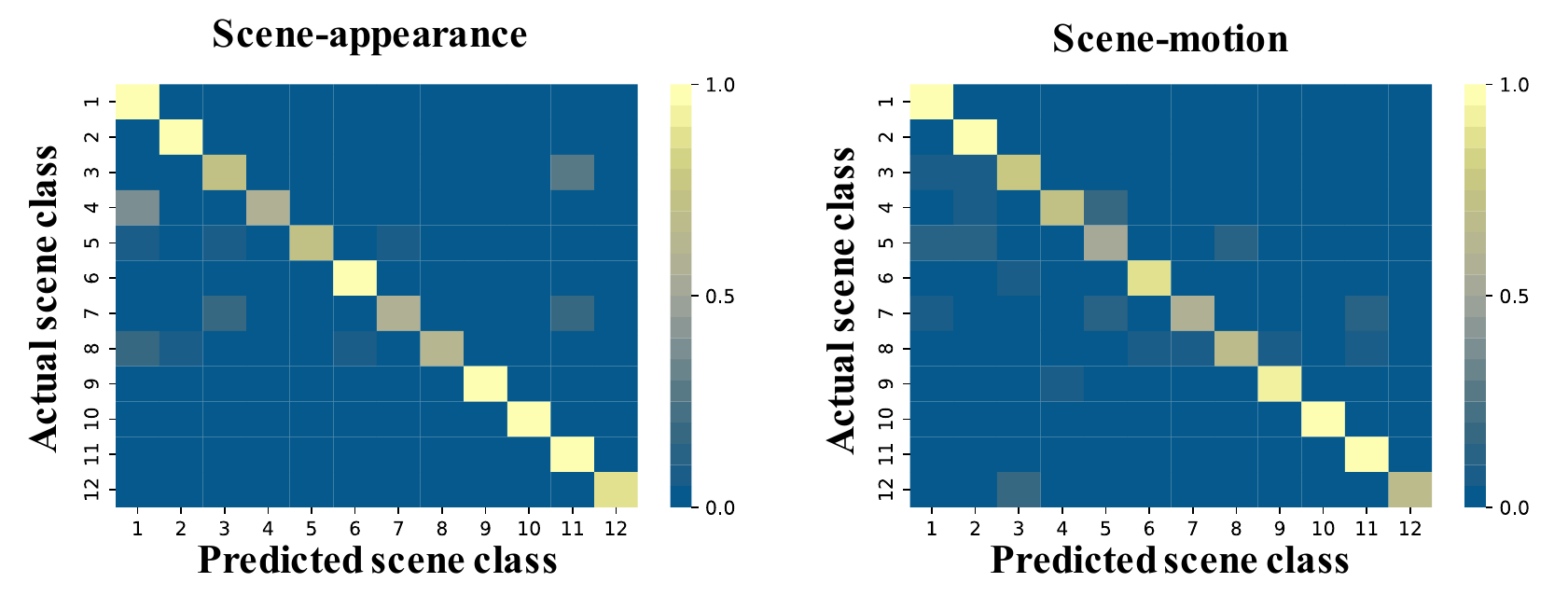}
	\caption{The confusion matrices of encoded scene-appearance and scene-motion features.  Best viewed in color.}
	\label{fig:scene_matching}
	\vspace{-14pt}
\end{figure}

\noindent\textbf{Impact of the memory size at test time.} The memories in our work are used for hierarchical semantic contrast during training and feature reconstruction at test time. In order to make our model more compact and efficient for inference, we may reduce the memory size by reserving a small portion of normal patterns. In this experiment, we randomly select a number of entries and discard the remaining at test time. ~\cref{fig:ablation_memory_size} illustrates the performance varies with the memory size. It shows that the performance is maintained well even if only 500 entries are reserved, and the performance only degenerates a little bit when only 100 entries are kept.

\begin{table}[tbp]
	\centering
	\caption{The AUC(\%) performance of more detailed variants on CUHK Avenue (Avenue), ShanghaiTech (SHT), and the scene-dependent mixture datasets (\ie [01,02] and [04,08]). When SA-E is not checked, only appearance/motion features are input to the encoders. When OC-D is not checked, the decoders reconstruct both scene and appearance/motion features.}
	\resizebox{.48\textwidth}{!}{
		\begin{tabular}{ccccc|cccc}
			\toprule
			SA-E & OC-D & $\mathcal{L}_{Scn}$ & $\mathcal{L}_{Obj}$ & $\mathcal{L}_{LC}$  & Avenue & SHT & [01,02] & [04,08]\\
			\midrule
			~ & \checkmark &   & ~ &   & 91.0 & 80.8 & 78.6 & 76.4\\
			\checkmark & ~ & ~ & ~ & ~ & 90.9 & 80.2 & 80.5 & 77.9 \\
			\checkmark & \checkmark & ~ & ~ & ~ & 91.1 & 80.7 & 81.0 & 78.2\\
			\midrule
			\checkmark & \checkmark & \checkmark & ~ & ~ & 91.3 & 81.6 & 82.1 & 79.0 \\
			\checkmark & \checkmark & ~ & \checkmark & ~ & 91.8 & 82.0 & 81.6 & 78.7\\
			\checkmark & \checkmark & \checkmark & \checkmark & ~ & 91.9 & 82.2 & 82.5 & 79.4\\
			\checkmark & \checkmark & \checkmark & ~ & \checkmark & 91.6 & 81.8 & 82.3 & 79.3\\
			\checkmark & \checkmark & ~ & \checkmark & \checkmark & 92.2 & 82.4 & 81.8 & 78.9\\
			\checkmark & \checkmark & \checkmark & \checkmark & \checkmark & \textbf{92.4} & \textbf{83.0} & \textbf{82.8} & \textbf{80.0}\\
			\bottomrule
		\end{tabular}
	}
	\label{tab:ablation_contrast_module}
	\vspace{-15pt}
\end{table}

\subsection{Visualization}
To investigate how well the hierarchical semantic contrast strategy works, we further analyze the scene classification results and the distribution of encoded latent features for data on ShanghaiTech. 

\noindent\textbf{The confusion matrix of scene classification.} We first investigate whether the encoded scene-aware features correctly fall into the actual scene clusters they belong to. To this end, we check the confusion matrix of scene classification for all test samples on ShanghaiTech, which contains 12 scenes. \cref{fig:scene_matching} (a) and (b) visualize the confusion matrices of encoded scene-appearance and scene-motion features, respectively. We observe that most encoded scene-aware features are correctly grouped.

\noindent\textbf{The distribution of encoded latent features.} We further investigate the distribution of encoded scene-aware normal features stored in the memory banks. \cref{fig:visual_inter_clustering} visualizes the distribution of them in the latent space, obtained by the models without or with HSC. We observe that the features distribute more compactly within classes and more separately between classes, consequently helping to discriminate anomalies from these normal patterns.

\subsection{Comparison to State-of-the-Art}
\label{sec:sota}
Finally, we compare our method with state-of-the-art. The comparison is first made on three public datasets which barely contain scene-dependent anomalies. To validate the effectiveness of our method on scene-dependent anomaly detection, we additionally make a comparison on the mixture datasets created upon ShanghaiTech.

\begin{table}[t]
	\centering
	\caption{Comparison results on UCSD Ped2 (Ped2), CUHK Avenue (Avenue), and ShanghaiTech (SHT). Besides the frame-level micro-averaged AUC(\%) performance, we also list the inputs of the methods, in which `F' denotes the frame-level input and `O' is object-centric. The subscript `A' is appearance, `F' is optical flow, `S' is skeleton, and `M' is other motion information. Besides, in our HSC model, $\text{MA}^{-,+}$ denotes using motion augmentation to generate both normal and abnormal samples.}
\resizebox{.48\textwidth}{!}{
	\begin{tabular}{c|c|c|c|c|c}
		\toprule
		Method & Reference & Input & Ped2 & Avenue & SHT \\
		\midrule
		AMC~\cite{nguyen2019anomaly} & ICCV19 & F & 96.2 & 86.9 & - \\
		Mem-AE~\cite{gong2019memorizing} & ICCV19 & F & 94.1 & 83.3 & 71.2 \\
		DeepOC~\cite{wu2019deepoc} & TNNLS19 & F & 96.9 & 86.6 & - \\
		r-GAN~\cite{lu2020fewshot} & ECCV20 & F & 96.2 & 85.8 & 77.9 \\
		CDAE~\cite{chang2020clustering} & ECCV20 & F & 96.5 & 86.0 & 73.3 \\
		MNAD~\cite{park2020learning} & CVPR20 & F & 97.0 & 88.5 & 72.8 \\
		IPR~\cite{tang2020integrating} & PRL20 & F & 96.3 & 85.1 & 73.0 \\
		LDF~\cite{ramachandra2020learning} & WACV20 & F & 94.0 & 87.2 & - \\
		CAC~\cite{wang2020cluster} & MM20 & F & - & 87.0 & 79.3 \\
		CT-D2GAN~\cite{feng2021D2GAN} & MM21 & F & 97.2 & 85.9 & 77.7 \\
		AMMCN~\cite{cai2021appearance} & AAAI21 & F & 96.6 & 86.6 & 73.7 \\
		MPN~\cite{lv2021learning} & CVPR21 & F & 96.9 & 89.5 & 73.8 \\
		AEP~\cite{yu2021abnormal} & TNNLS21 & F & 97.9 & 90.2 & - \\
		SIGnet~\cite{fang2022SIGnet} & TNNLS22 & F & 96.2 & 86.8 & - \\
		IAAN~\cite{zhang2022influence} & TCSVT22 & F & 92.9 & 80.5 & 80.3 \\
		ROADMAP~\cite{wang2022ROADMAP} & TNNLS22 & F & 96.4 & 88.3 & 76.6 \\
		\midrule
		GEPC~\cite{markovitz2020graph} & CVPR20 & $O_S$ & - & - & 76.1 \\
		STGformer~\cite{huang2022hierarchical} & MM22 & $O_S$ & - & 88.8 & 82.9 \\
		\midrule
		HSNBM~\cite{bao2022scene} & MM22 & F+$O_A$ & 95.2 & 91.6 & 76.5 \\
		\midrule
		STC-Graph~\cite{sun2020scene} & MM20 & $O_A$+$O_M$ & - & 89.6 & 74.7 \\
		SSMTL\footnotemark[1]~\cite{georgescu2021anomaly} & CVPR21 & $O_A$+$O_M$ & 97.5 & 91.5 & 82.4 \\
		\midrule
		VEC~\cite{yu2020cloze} & MM20 & $O_A$+$O_{F}$ & 97.3 & 90.2 & 74.8 \\
		$\text{HF}^2$-VAD~\cite{liu2021hybrid} & ICCV21 & $O_A$+$O_{F}$ & \textbf{99.3} & 91.1 & 76.2 \\
		BAF~\cite{Georgescu2022AED} & TPAMI22 & $O_A$+$O_{F}$ & 98.7 & 92.3 & 82.7 \\
		CAFE~\cite{yu2022effective} & MM22 & $O_A$+$O_{F}$ & 98.4 & 92.6 & 77.0 \\
		DERN~\cite{sun2022evidential} & MM22 & $O_A$+$O_{F}$ & 97.1 & 92.7 & 79.3 \\
		BDPN~\cite{chen2022comprehensive} & AAAI22 & $O_A$+$O_{F}$ & 98.3 & 90.3 & 78.1 \\
		\midrule
		HSC & \multirow{2}*{This work} & \multirow{2}*{$O_A$+$O_S$} & \multirow{2}*{$\text{98.1}^{\dagger}$} & 92.4 & 83.0\\
		HSC w/ $\text{MA}^{-,+}$ &  &  &  & \textbf{93.7} & \textbf{83.4} \\
		\bottomrule
	\end{tabular}
}
\label{tab:scene_agnostic_results}
\vspace{-12pt}
\end{table}

\footnotetext[1]{Here the micro-average AUC is reported from the officially released website https://github.com/lilygeorgescu/AED-SSMTL.}

\noindent\textbf{Comparison on scene-independent VAD.} We compare our method with recent VAD methods that learn from normal data as well. The comparison results are presented in \Cref{tab:scene_agnostic_results}, in which the inputs of all methods are also provided for reference. Generally speaking, benefiting from the extracted high-level features, most of the object-centric methods perform better than the methods with frame-level inputs, although some of the latter methods also use motion information like optical flow. In addition, the proposed method outperforms all other methods on both Avenue and ShanghaiTech, validating the superiority of our design. Note that we are not able to test our full model on the UCSD Ped2 dataset due to its low resolution, in which no high-quality skeleton keypoints can be detected. Therefore, we only use the scene-appearance stream of our model for testing, which still achieves a performance higher than a great number of methods. 

\noindent\textbf{Comparison on scene-dependent VAD.} Finally, we investigate the performance of our method on scene-dependent anomaly detection based on the mixture datasets introduced above. The results are presented in \Cref{tab:mixture_auc}. We also test other SOTA methods~\cite{gong2019memorizing,park2020learning,lv2021learning,liu2021hybrid} using their released codes for comparison. Unfortunately, we are not able to compare with the scene-aware methods~\cite{sun2020scene,bao2022scene,chen2022comprehensive} since their codes are not available. The results show that the performance of the other methods, especially those with frame-level inputs, degenerates dramatically. In contrast, all model variants of our proposed method consistently demonstrate promising performance.

\textsc{\begin{table}[t]
  \centering
  \caption{Comparison results on the scene-dependent mixture datasets built upon ShanghaiTech. $\text{MA}^{-,+}$ denotes using motion augmentation to generate both normal and abnormal samples, $\text{MA}^{-}$ denotes augmenting normal samples only.}
  \resizebox{.46\textwidth}{!}{
    \begin{tabular}{c|c|c|c|c|c}
      \toprule
      Method & Reference & Input & [01,02] & [04,08] & [10,12] \\
      \midrule
      Mem-AE~\cite{gong2019memorizing} & ICCV19 & F & 77.7 & 60.2 & 50.2 \\
      MNAD~\cite{park2020learning} & CVPR20 & F & 77.8 & 68.6 & 50.0 \\
      MPN~\cite{lv2021learning} & CVPR21 & F & 78.4 & 61.5 & 45.3 \\
      $\text{HF}^2$-VAD~\cite{liu2021hybrid} & ICCV21 & $O_A$+$O_F$ & 74.8 & 75.2 & 66.8 \\
      \midrule
      HSC & \multirow{3}*{This work} & \multirow{3}*{$O_A$+$O_S$} & 82.8 & 80.0 & 87.3 \\
      HSC w/ $\text{MA}^{-}$ &  & & 85.7 & 81.8 & 90.1 \\
      HSC w/ $\text{MA}^{-,+}$ &  & & \textbf{86.9} & \textbf{82.6} & \textbf{91.0} \\
      \bottomrule
    \end{tabular}
  }
  \label{tab:mixture_auc}
  \vspace{-12pt}
\end{table}}

\vspace{-5pt}
\subsection{Limitations}
Since our proposed method takes skeleton-based motion features as one of the inputs, the full model is restricted to human-related datasets and requires the datasets are not very low in resolution. Otherwise, only the scene-appearance stream works, which inevitably degenerates the performance. A possible extension is replacing the skeleton-based motion features with optical flow and conducting contrastive learning based on the clustering results of optical flow features. Besides, other components of this framework can be replaced by other advanced modules, \eg substituting another advanced background parsing model for the simple segmentation map.

\section{Conclusion}
In this work, we have presented a hierarchical semantic contrast method to address scene-dependent video anomaly detection. The design of our hierarchical semantic contrastive learning, together with scene-aware autoencoders and motion augmentation, enables the proposed model to achieve promising results on both scene-independent and scene-dependent VAD. Experiments on three public datasets and self-created datasets have validated the effectiveness of our method.

{\small
\bibliographystyle{ieee_fullname}
\bibliography{egbib}
}

\end{document}